\documentclass[10pt,twocolumn,letterpaper]{article}

\usepackage[table,xcdraw]{xcolor}
\usepackage{amssymb}
\usepackage{multirow}
\usepackage[pagenumbers]{cvpr} 

\definecolor{cvprblue}{rgb}{0.21,0.49,0.74}
\usepackage[pagebackref,breaklinks,colorlinks,citecolor=cvprblue]{hyperref}

\def\paperID{5851} 
\def\confName{CVPR}
\def\confYear{2024}

\title{Clean Label Disentangling for Medical Image Segmentation with Noisy Labels}

\author{Zicheng Wang~~~~Zhen Zhao~~~~Erjian Guo~~~~Luping Zhou \\
}

\begin{document}
\maketitle

\begin{abstract}
Current methods focusing on medical image segmentation suffer from incorrect annotations, which is known as the noisy label issue. Most medical image segmentation with noisy labels methods utilize either noise transition matrix, noise-robust loss functions or pseudo-labeling methods, while none of the current research focuses on clean label disentanglement. We argue that the main reason is that the severe class-imbalanced issue will lead to the inaccuracy of the selected ``clean'' labels, thus influencing the robustness of the model against the noises. In this work, we come up with a simple but efficient class-balanced sampling strategy to tackle the class-imbalanced problem, which enables our newly proposed clean label disentangling framework to successfully select clean labels from the given label sets and encourages the model to learn from the correct annotations. However, such a method will filter out too many annotations which may also contain useful information. Therefore, we further extend our clean label disentangling framework to a new noisy feature-aided clean label disentangling framework, which takes the full annotations into utilization to learn more semantics. Extensive experiments have validated the effectiveness of our methods, where our methods achieve new state-of-the-art performance. Our code is available at https://github.com/xiaoyao3302/2BDenoise.
\end{abstract}    
\section{Introduction}
\label{sec_introduction}

Medical image segmentation is an urgent vision task that contributes to medical image reasoning, which is vital in the development of the computer-aided diagnosis (CAD) system~\cite{valanarasu2022unext, yan2022after, yao2022enhancing, wu2022mutual, cao2022swin}. Current medical image segmentation methods rely heavily on deep neural networks, which require extensive fully pixel-level annotated data for the model training~\cite{wang2023conflict, zhao2023instance, zhao2023augmentation}. However, the boundary region of medical images is hard to distinguish. Therefore, it introduces great difficulty to collect a large precisely annotated medical image dataset~\cite{bai2023bidirectional, wu2022exploring, luo2022semi, yao2022enhancing}. Consequently, the deep models on medical images are far from reaching their potential. Currently, various works turn to using less accurately labeled data to train deep models to reduce the annotation pressure~\cite{zhang2022cyclemix, luo2022scribble, chen2022c}. These less accurate labels can be obtained through coarse annotation, or generated by applying a model trained on other public datasets on a new dataset, etc. However, such a training strategy will inevitably lead to the training being misled by the wrong annotations, which is also known as the noisy label issue~\cite{guo2021metacorrection, guo2022simt, li2021superpixel, han2018co}. 

\begin{figure}[t]
\centering
\includegraphics[width=0.9\linewidth]{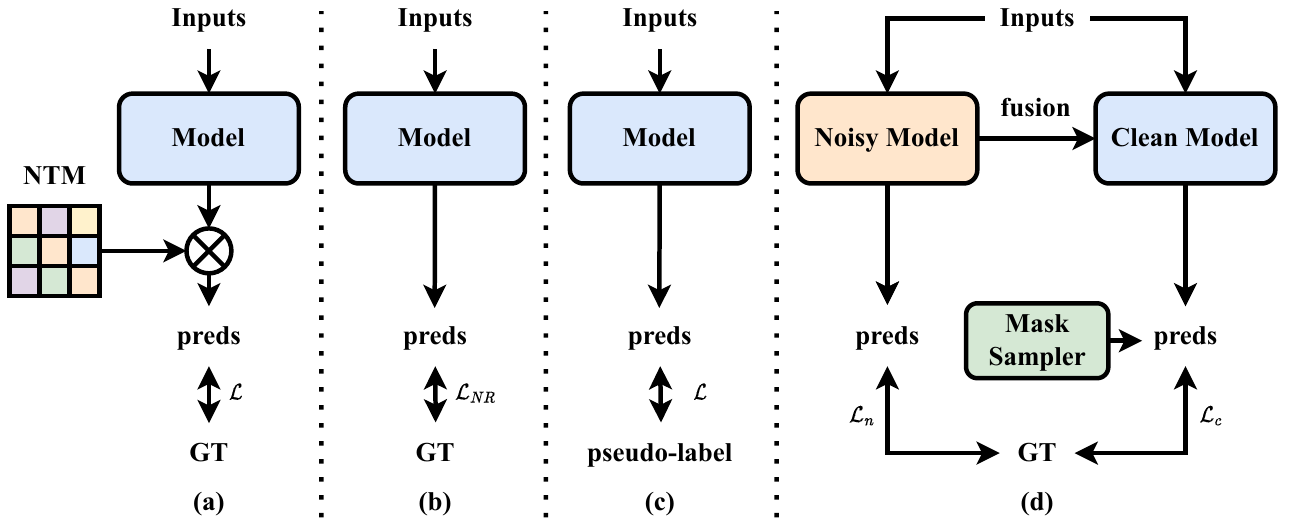}
\caption{An overview of the pipelines of different methods. (a) indicates the pipeline of the methods that focus on estimating the noise transition matrix to handle the noisy labels. (b) indicates the pipeline of the methods that propose noise-robust loss functions to mitigate the influence of the noisy labels. (c) indicates the pipeline of the methods that generate pseudo-labels that are more reliable than the noisy labels to train the model. (d) indicates the pipeline of our newly proposed noisy feature-aided clean label disentangling (NA-CLD) framework.}
\vspace{-5mm}
\label{fig_different_pipeline}
\end{figure}

Facing the issue, recent research focus on designing new training strategies to mitigate the influence of noisy labels. Current research about deep medical image segmentation with noisy labels (MISNL) can be divided into three main categories. 
The first category includes the methods that focus on estimating the noise transition matrix to handle the noisy labels~\cite{guo2022joint, li2021provably}, as shown in Fig.~\ref{fig_different_pipeline} (a). However, these methods can only estimate the class-level transition matrix, which is not so effective for randomly distributed noises. 
The second category includes the methods that propose noise-robust loss functions to mitigate the influence of the noisy labels~\cite{wang2020noise, ghosh2017robust, wang2019symmetric}, as shown in Fig.~\ref{fig_different_pipeline} (b). However, these loss functions are only partially robust to noisy labels, leading to an unstable performance. 
The third category includes the methods that propose to enable the model to generate reliable pseudo-labels and use the pseudo-labels to train the model~\cite{zhang2020robust, yao2022learning}, as shown in Fig.~\ref{fig_different_pipeline} (c). However, these methods may suffer from the confirmation bias issue.

In the meanwhile, there is no MISNL method focused on clean label disentanglement so far, which is commonly used in other domains~\cite{liu2022perturbed, wang2022semi, yang2023revisiting}. Based on our observations, we argue that the main reason is the severe class-imbalanced problem, which is commonly seen in the medical image domain~\cite{you2023bootstrapping}. The clean label disentanglement will filter out less confident predictions, which based on our observation may most lie in the minor categories, and the remained annotations may probably lie in the major categories. Therefore, the training will be biased by the class-imbalanced problem, leading to the degradation of the recognition performance of the model~\cite{guan2022unbiased}.

Facing the above-mentioned problems, in this paper, we propose a new noisy feature-aided clean label disentangling (NA-CLD) framework to tackle the MISNL problem, as listed in Fig.~\ref{fig_different_pipeline} (d), which aims at encouraging the model to learn from clean annotations, thus reducing the influence of the noisy labels. 

\begin{figure}[t]
\centering
\includegraphics[width=0.9\linewidth]{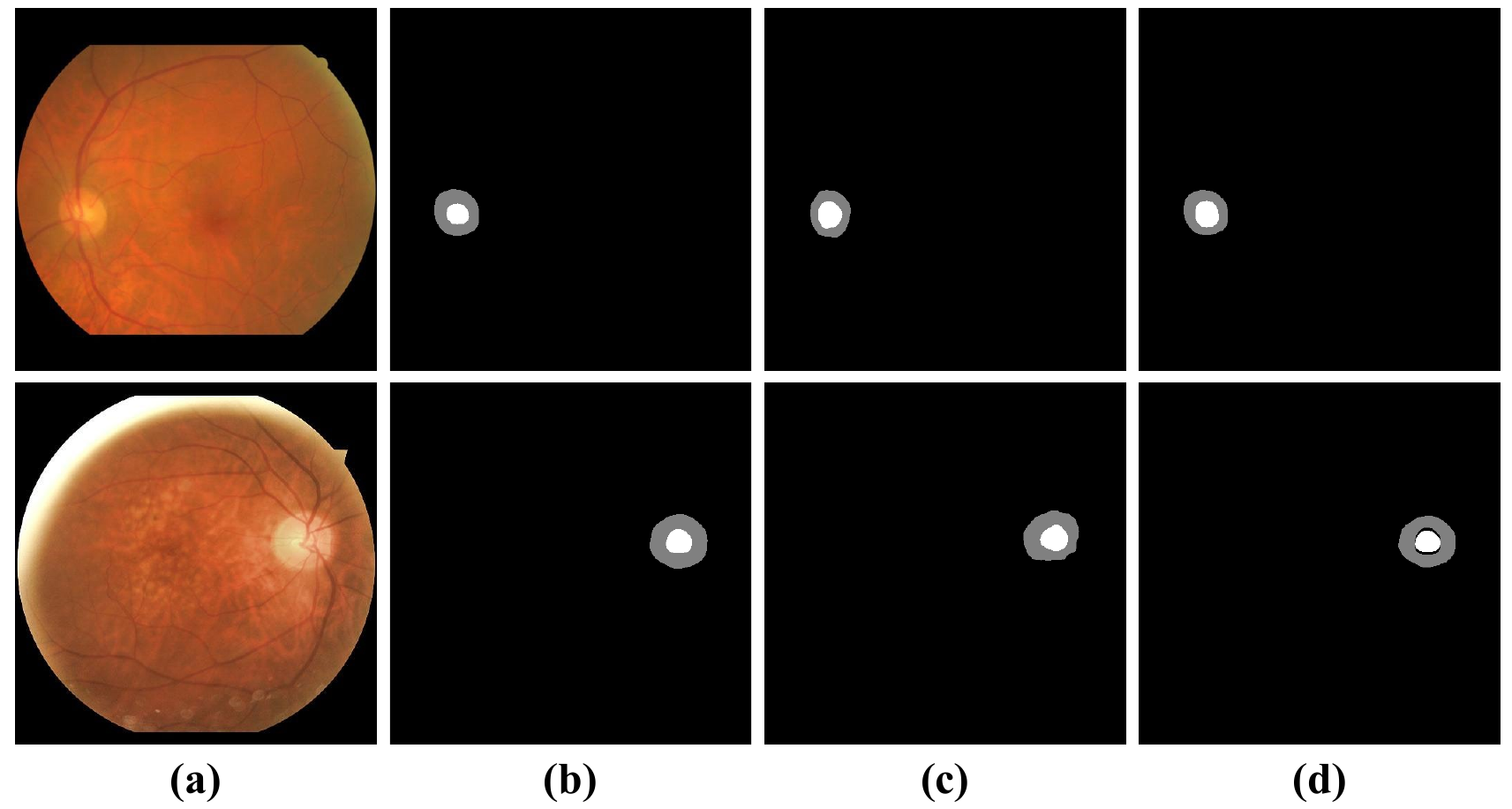}
\caption{The visualization of different kinds of label noises. (a) the input images, (b) the ground-truth labels, (c) the source-free domain adaptation (SFDA) noise, and (d) the joint erode and dilation (JED) noise. The samples are collected from the G1020 dataset.}
\vspace{-5mm}
\label{fig_noise}
\end{figure}

In particular, we first come up with a clean label disentangling (CLD) framework to filter out clean labels from the given noisy labels according to the prediction confidence of the model, and then we encourage the model to learn semantics from these clean annotations. 
However, as our CLD framework will filter some annotations, it will make the class-imbalanced problem even more severe and degrade the recognition performance of the model. To tackle the problem, we further propose a class-balanced sampling (CBS) strategy to adaptively re-sample the filtered clean annotations according to the number of filtered annotations belonging to different categories, in this way, we can make sure that the sampled annotations are distributed evenly to reduce the influence of the class-imbalanced problem, thus enabling the model to learn unbiased semantics. Our extensive experiments have validated the necessity of dealing with the class-imbalanced problem to the noisy label issue.

Nevertheless, our CLD and CBS strategies both may filter out too many annotations that can also contain considerable useful information~\cite{aimar2023balanced, kim2020m2m}. Therefore, we further come up with a noisy feature-aided clean label disentangling (NA-CLD) framework to jointly tackle the limitations of these two strategies, where we introduce an extra noisy encoder to extract features from the initial noisy label sets, and the features extracted are then fused with the clean features extracted by the clean encoder in our CLD framework to help the clean model to learn more potential information from the filtered samples to generate more accurate predictions.

We adopt two types of noisy labels, \textit{i.e.}, the source-free domain adaptation (SFDA) noise~\cite{guo2022joint} and the joint erode and dilation (JED) noise~\cite{li2021superpixel}, to verify that our method is robust to different noise types, where the SFDA noise is constructed by generating pseudo-labels for a given dataset using a model trained on another public dataset, and the JED noise is constructed by randomly erode or dilate the coverage area of a certain category, as shown in Fig.~\ref{fig_noise}. We have validated the effectiveness of our NA-CLD method on various benchmark datasets,\textit{i.e.}, the G1020 fundus image dataset~\cite{bajwa2020g1020} and the breast ultrasound image dataset B~\cite{yap2017automated}, where our method achieves new state-of-the-art (SOTA) performance.

Our contributions can be summarized in four folds:

\begin{enumerate}
    \item We propose a new clean label disentangling (CLD) framework to enable the model to learn semantics from the filtered correct annotations.
    \item We propose a class-balanced sampling (CBS) strategy to tackle the class-imbalanced issue and enable the model to learn unbiased semantics.
    \item We propose a new noisy feature-aided clean label disentangling (NA-CLD) framework to enable the model to learn more semantics from both filtered correct annotations and filtered-out annotations.
    \item Our NA-CLD method achieves state-of-the-art (SOTA) performance on benchmark datasets.
\end{enumerate}
\section{Related work}
\label{sec_related}
In this section, we briefly discuss deep learning for medical image segmentation and deep segmentation methods with noisy labels, which are related to our work.

\subsection{Deep learning for medical image segmentation}


Most state-of-the-art medical image segmentation methods are based on deep neural networks adopted from the natural image domain like the DeepLab series~\cite{chen2014semantic, chen2017deeplab, chen2017rethinking, chen2018encoder} and the Transformer series~\cite{vaswani2017attention, dosovitskiy2020image, carion2020end}, \emph{etc}. However, most of these deep models are not carefully designed for medical image segmentation tasks. Different from natural images, medical images have some special properties like the extreme class-imbalanced problem. These properties often lead to poor performance of deep models designed for natural images when directly applied to medical images~\cite{liu2023samm, wu2023medical}.
Various works focusing on medical image analysis have been proposed to tackle the problem by proposing different kinds of loss functions like Dice loss~\cite{milletari2016v}, Combo loss~\cite{taghanaki2019combo}, etc. However, these loss functions are sensitive to hyper-parameters which need to be carefully adjusted when applied to a new scenario.

\subsection{Deep segmentation methods with noisy labels}
In the natural image domain, most research have been focused on tackling the noisy labels issue on the classification tasks~\cite{liu2015classification, xia2019anchor, xia2020robust}. However, compared with natural images, the study of medical images may focus more on image reasoning like semantic segmentation~\cite{shen2017deep}, while only a few works have been focused on medical image segmentation with noisy labels (MISNL) problems. 

Current research about deep segmentation with noisy labels can be divided into three main categories. 

The first category includes the methods that focus on estimating the noise transition matrix~\cite{li2021provably, guo2022joint}. 
VolMin~\cite{li2021provably} proposes an end-to-end framework to jointly optimize the transition matrix and the classifier, which reduces the reliance of the model on the anchor points.
JCAS~\cite{guo2022joint} considers both pixel-wise label correction and pair-wise label correction to handle the label noise heterogeneously. 
However, these methods can only estimate the class-level transition matrix, which is not so effective for randomly distributed noises. In addition, we observe that these methods cannot handle the class-imbalanced problem well, which may also degrade the recognition performance of the model.

\begin{figure*}[t]
\centering
\includegraphics[width=\linewidth]{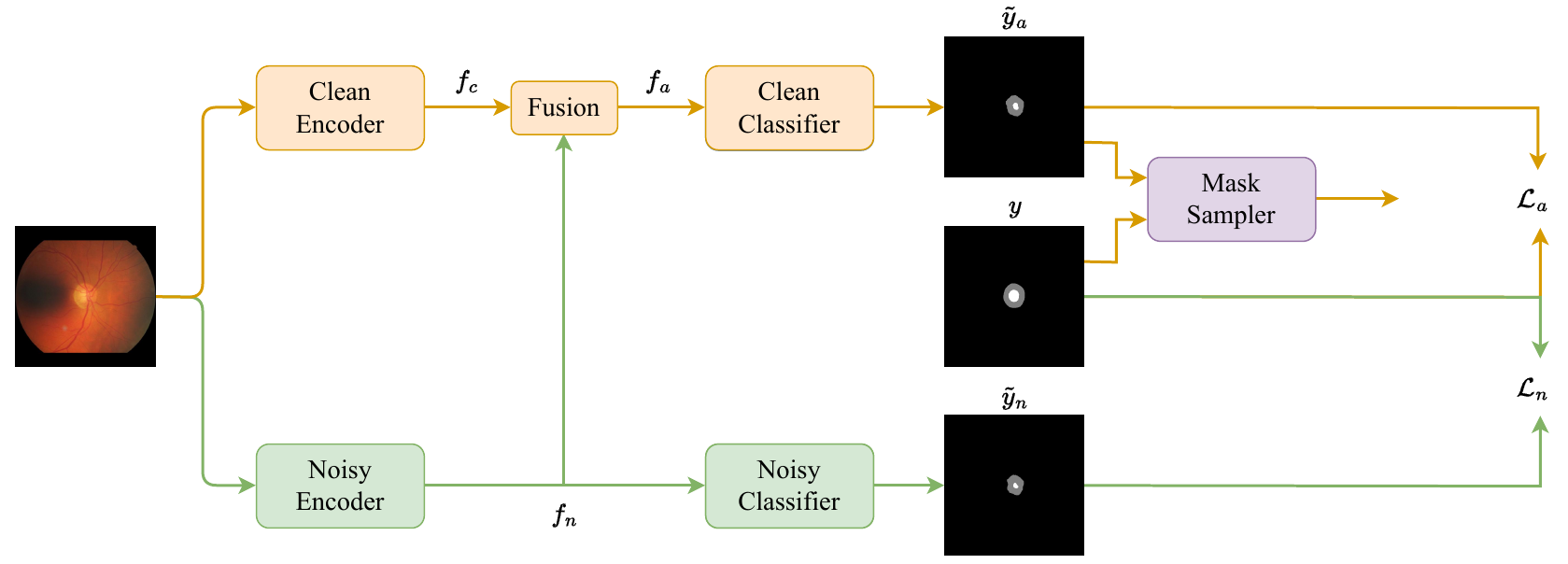}
\caption{The network architecture of our noisy feature-aided clean label disentangling (NA-CLD) framework. On one hand, we use a mask sampler to disentangle clean labels from the given noisy label set and use the clean labels to train a clean model, including a clean encoder to extract clean features from the given images and a clean classifier to generate predictions given the input features. On the other hand, we further use the initial label set to learn a noisy model, including a noisy encoder to extract noisy features from the given images and a noisy classifier to generate predictions given the input noisy features. In addition, considering that our clean label disentangling strategy will drop out too many annotations, we further use a fusion module to enable the noisy feature, which may also contain useful information, to aid the clean model learning to generate more accurate predictions.}
\label{fig_NA_CLD}
\end{figure*}

The second category includes the methods that proposed noise-robust loss functions~\cite{wang2020noise, ghosh2017robust, wang2019symmetric}.
MAE~\cite{ghosh2017robust} validates that loss functions that are based on the mean absolute value of error are robust enough to the label noise. 
SCE loss~\cite{wang2019symmetric} combines a Cross-Entropy (CE) loss and a noise-robust Reverse Cross Entropy (RCE) loss, which can not only contribute to hard category learning but also be robust to noisy labels.
However, such loss functions cannot handle the class-imbalanced problem efficiently. Facing the issue, NR-Dice~\cite{wang2020noise} combines the MAE loss and the Dice loss and proposes a noise-robust Dice loss. NR-Dice loss treats different pixels differently, which is robust to noisy labels and can also handle the class-imbalanced problem. However, such a loss is sensitive to the hyper-parameters.
Moreover, these noise-robust loss functions are only partially robust to noisy labels.

\begin{figure}[t]
\centering
\includegraphics[width=\linewidth]{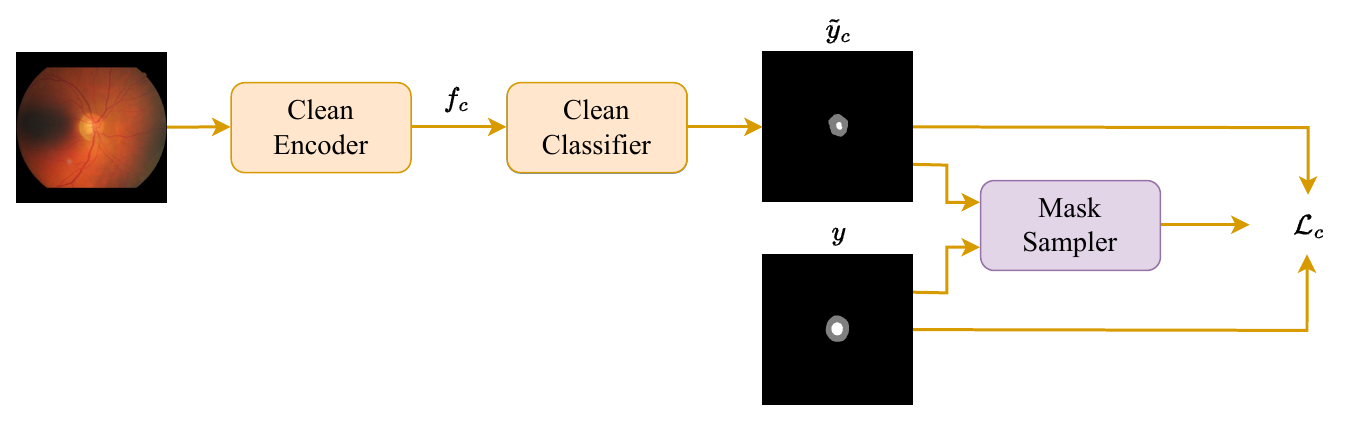}
\caption{The network architecture of our clean label disentangling (CLD) framework where we use a mask sampler to filter out clean annotations to learn a clean model.}
\vspace{-3mm}
\label{fig_CLD}
\end{figure}

The third category includes the methods that aim at generating pseudo-labels for model training, while the pseudo-labels might be more reliable than the given noisy labels~\cite{zhang2020robust, yao2022learning}.
Tri-Network~\cite{zhang2020robust} extends the co-teaching framework by proposing a tri-network and enables two sub-nets to generate confident pseudo-labels for the third sub-net. Therefore, the model will benefit from the generated high-quality annotations.
SC~\cite{yao2022learning} proposed a new Markov label noise and propose an iterative training process by enabling the model to generate pseudo-labels for self-training.
However, these methods may suffer from the confirmation bias issue and the training is not stable.

Moreover, none of the previous works that focus on MISNL take the class-imbalanced problem into consideration, which may also severely degrade the recognition performance of the model and affect the performance of some noise robust or noise correction methods. Therefore, in this work, we have validated the necessity of dealing with the class-imbalanced problem to the noisy label issue.
\section{Methodology}
\label{sec_method}
In this section, we will introduce our newly proposed noisy feature-aided clean label disentangling (NF-CLD) framework in detail. In Sec.~\ref{sec_problem} we will give a brief definition of the problem of medical image segmentation with noisy labels (MISNL). Then, we will introduce our clean label disentangling (CLD) framework in Sec.~\ref{sec_CLD}, our class-balanced sampling (CBS) strategy in Sec.~\ref{sec_CBS} and our noisy feature-aided clean label disentangling (NF-CLD) framework in Sec.~\ref{sec_NACFL}.

\subsection{Medical Image Segmentation with Noisy Labels}
\label{sec_problem}
In the MISNL task, we are given a set of medical images $\mathcal{X}=\{x_i\}_{i=1}^{M}$ and the corresponding annotations $\mathcal{Y}=\{y_i\}_{i=1}^{M}$. The $x_i \subset \mathbb{R}^{H\times W\times C}$ indicates the input image with a size of $H\times W$ and $C$ channels, while $y_i \subset \left\{ 0, 1\right\}^{H\times W\times Y}$ is the one-hot ground truth label for each pixel, where $Y$ indicates the number of visual classes in total. Note that the annotations in the training stage are accompanied by noise, \textit{i.e.}, some of the annotations are incorrect, while the clean annotations are only available during the inference stage for validating the performance.

\subsection{Clean Label Disentangling Framework}
\label{sec_CLD}

In this section, we illustrate our newly proposed clean label disentangling (CLD) framework in detail. The pipeline of our CLD is shown in Fig.\ref{fig_CLD}, where we aim at training a clean model $\Psi_c$ with the disentangled clean annotations. Here we divide the clean model $\Psi_c$ into a clean encoder $\Psi_{f, c}$, which aims at extracting features $f_c$ from the input images $x_i$, and a clean classifier $\Psi_{cls, c}$, which aims at generating pixel-wise predictions $\widetilde{y}_{c}$ using the extracted features $f_c$. Similarly, we also denote the predicted category of a prediction as $\hat{y}_{c}$ and the confidence score of a prediction as $\check{y}_{c}$. Recall that we aim at disentangling clean labels from the given noisy label set, we thereby come up with a mask sampler to sample out clean masks for the clean annotations. However, which annotations are clean is unknown. Therefore, we propose a voting mechanism to utilize the predictions of the model and the given label sets to distinguish which labels are clean. Here we use a binary variable $\hat{\delta}_{k, i}$ to denote the mask of the annotations, which indicates the mask for the $k$-th pixel of the $i$-th annotation, where $\hat{\delta}_{k, i}$ equals 0 or 1, indicating whether the annotation is treated as the noisy label of the clean label, respectively. We come up with two different voting strategies for clean mask sampling, \textit{i.e.}:

\textbf{Strategy 1:} If $\check{y}_{c}^{k, i}$ is larger than a pre-determined confidence threshold $\gamma$, then we will set $\hat{\delta}_{k, i}$ as 1, otherwise we will set $\hat{\delta}_{k, i}$ as 0.

\textbf{Strategy 2:} If $\check{y}_{c}^{k, i}$ is larger than a pre-determined confidence threshold $\gamma$ \textbf{or} $\hat{y}_{c}^{k, i} = y_{c}^{k, i}$, then we will set $\hat{\delta}_{k, i}$ as 1, otherwise we will set $\hat{\delta}_{k, i}$ as 0.

Our Strategy 1 relies on the predictions of the model and aims at enabling the model to generate pseudo-labels as supervision for self-training. However, such an operation relies heavily on the recognition performance of the model, which may lead to the training being infected by the confirmation bias problem.

To tackle the issue, we further propose Strategy 2, which is based on a voting mechanism. We argue that both the prediction of the model and the given noisy labels can contain useful information, and the prediction of the noise-robust model is more reliable. Therefore, we come up with a voting strategy that if the prediction score $\check{y}_{c}^{k, i}$ is higher than the predefined threshold, we prefer to treat the prediction of the model to be the clean label. Otherwise, if the prediction of the model is equal to the given annotation, we can also treat the annotation as correct, as the probability of the two indicators being wrong at the same time is small. In this way, those reliable annotations can be filtered out.

In addition, inspired by ADELE~\cite{liu2022adaptive} that the model would probably fit the clean data at the beginning stage of the training process, we adopt a disentangling after warm-up strategy by firstly training the model using all of the given annotations as supervision for a couple of epochs, \textit{i.e.}, $\tau$ epochs, and then perform our proposed clean label disentangling strategy. In this way, we can not only make sure that the model will generate semantic meaningful predictions to guarantee the accuracy of the voting, but can also prevent the training from being influenced by the noisy labels.

To sum up, when learning the model, we use the filtered clean annotations to supervise the clean model training, and the clean stream loss function $\mathcal{L}_{c}$ can be written as:
\begin{equation}\label{eq_loss_CLD}
    \mathcal{L}_{c} = \frac{\sum_{i=1}^{M} \sum_{k=1}^{W \times H} \ell_{ce}(\widetilde{y}_{c}^{k, i}, y_{c}^{k, i}) \times \hat{\delta}_{k, i}}{\sum_{i=1}^{M} \sum_{k=1}^{W \times H} \hat{\delta}_{k, i}}
\end{equation}

\subsection{Class-Balanced Sampling}
\label{sec_CBS}

\begin{table}[b]\centering
\footnotesize
\caption{Comparison of the distribution of the number of samples (in \%) contained in different categories (\textit{i.e.}, the Background, the Optic Disk and the Optic Cup) using or without using our proposed clean label disentangling (CLD) framework and class balance sampling (CBS) strategy. The samples are from the G1020 datasets with SFDA noise.}
\scalebox{0.85}{
\begin{tabular}{l|cccc}
\hline
\textbf{Method} & \textbf{Background (\%)} & \textbf{Optic Disk (\%)} & \textbf{Optic Cup (\%)} \\
\hline
\textbf{w/o CLD, w/o CBS} & 98.99 & 0.81 & 0.20 \\
\textbf{w/ CLD, w/o CBS} & 99.07 & 0.74 & 0.19 \\
\hline
\textbf{w/o CLD, w/ CBS} & 76.53 & 17.22 & 6.27 \\
\textbf{w/ CLD, w/ CBS} & 78.16 & 16.08 & 5.75 \\
\hline
\end{tabular}
}
\label{table_class_imbalance}
\end{table}

However, our clean label disentangling (CLD) framework relies heavily on the prediction of the model, while the class-imbalanced problem which is commonly seen in the medical image analysis domain would severely affect the performance of our CLD strategy, as the prediction scores of some categories may not be that high. Moreover, our CLD strategy may even exacerbate the class-imbalanced problem, as listed in Table~\ref{table_class_imbalance}, leading to a degradation of the recognition performance of the model. To tackle the problem, we further propose a class-balanced sampling (CBS) strategy to mitigate the long-tail effect.

\begin{figure}[h]
\centering
\includegraphics[width=\linewidth]{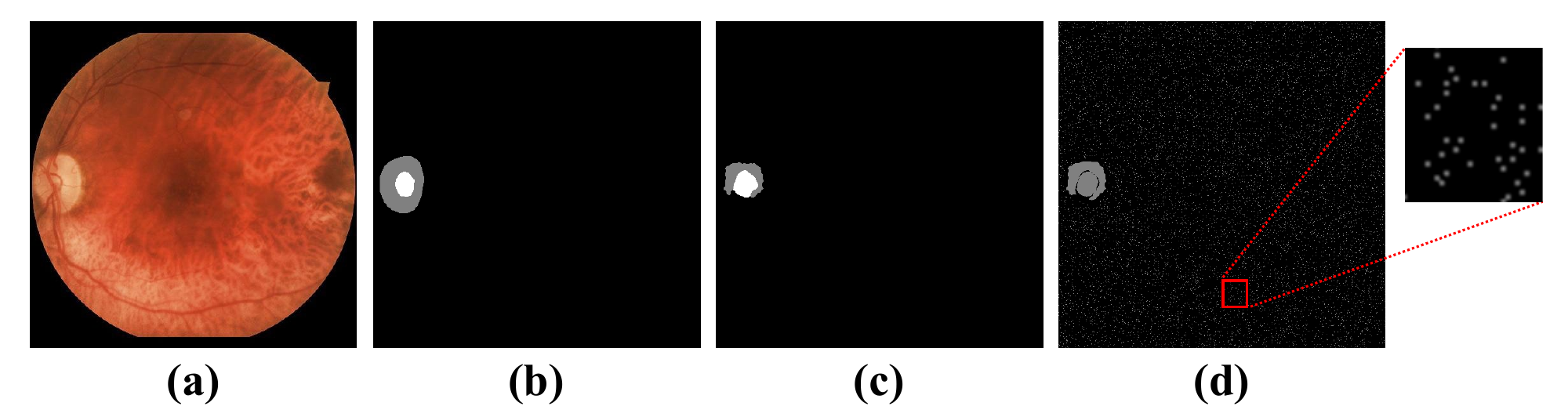}
\caption{The visualization of our class-balance sampling (CBS) strategy. (a) the input image, (b) the ground-truth label, (c) the source-free domain adaptation (SFDA) noise, and (d) the mask sampled by our CBS strategy, the gray region indicates the pixels where $\hat{\delta}_{k, i} = 1$.}
\label{fig_CBS}
\end{figure}

In particular, given a set of filtered clean annotations, we expect that the number of pixels occupied by the category with the most pixels is no more than $\rho$ times the number of pixels occupied by the category with the least pixels. Therefore, given the number of pixels $\omega$ occupied by the category with the least pixels, we perform a class-balanced sampling to sample out a sub-set from the category that occupies more than $\rho \cdot \omega$ pixels. Here we also use a binary variable $\check{\delta}_{k, i}$ to denote the class-balanced sampling mask of the annotations where $\check{\delta}_{k, i}$ equals 0 or 1, indicating whether the annotation is sampled out or remained, respectively. Considering the spatial position relationship of the sampled pixels, we adopt the random sampling strategy to perform the class-balanced sampling, which is shown in Fig.~\ref{fig_CBS}.

Therefore, when learning the model, we use the class-balance sampled clean annotations to supervise the training, and the clean stream loss function $\mathcal{L}_{c}$ can be written as:
\begin{equation}\label{eq_loss_CBS}
    \mathcal{L}_{c} = \frac{\sum_{i=1}^{M} \sum_{k=1}^{W \times H} \ell_{ce}(\widetilde{y}_{c}^{k, i}, y_{c}^{k, i}) \times \hat{\delta}_{k, i} \times \check{\delta}_{k, i}}{\sum_{i=1}^{M} \sum_{k=1}^{W \times H} \hat{\delta}_{k, i} \times \check{\delta}_{k, i}}
\end{equation}

It can be inferred from Table~\ref{table_class_imbalance} that when applied our CBS strategy, the class-imbalanced problem can be efficiently alleviated, leading to an unbiased training process.

\subsection{Noisy Feature-Aided Clean Label Disentangling}
\label{sec_NACFL}
Our CLD framework together with our CBS strategy can efficiently disentangle clean annotations from the given noisy label sets. However, such operations will filter out too many annotations which may also contain useful semantic information, as listed in Table~\ref{table_remain_annotation}. 

To tackle the problem, in this section, we further come up with a noisy feature assistance (NFA) strategy and propose a noisy feature-aided clean label disentangling (NF-CLD) framework, where we learn an additional noisy model, including a noisy encoder $\Psi_{f, n}$ and a noisy classifier $\Psi_{cls, n}$, that uses the initial noisy label set as supervision, and then enable the features extracted by the noisy encoder to aid the clean model learning. In particular, we use a channel-wise concatenation operation followed by a non-linear fusion layer $\Psi_{fuse}$ to fuse the noisy feature $f_n$ with the clean feature $f_c$ to obtain the noisy feature-aided clean feature (all feature), denoted as $f_a$. On one hand, the fusion module will improve the quality of the input features of the clean encoder, as $f_a$ will contain much more semantic information compared with $f_c$. On the other hand, the fusion module will also enable the class-balanced clean annotations to perform unbiased supervision on the noisy features, which will also contribute to the noisy feature extraction. Here we denote the noisy predictions generated by $\Psi_{cls, n}$ as $\widetilde{y}_{n}$, with the predicted category of a prediction as $\hat{y}_{n}$ and the confidence score of a prediction as $\check{y}_{n}$. The noisy stream loss function $\mathcal{L}_{n}$ can be written as:
\begin{equation}\label{eq_sup_single}
    \mathcal{L}_{n} = \frac{1}{M} \sum_{m=1}^{M} \frac{1}{W\times H} \sum_{k=1}^K \ell_{ce}(\widetilde{y}_{n}^{k, i}, y_{n}^{k, i})
\end{equation}

\begin{table}[t]\centering
\footnotesize
\caption{Comparison of the number of the remained annotations (in \%) using or without using our proposed clean label disentangling (CLD) framework and class balance sampling (CBS) strategy. The samples are from the G1020 datasets with SFDA noise.}
\scalebox{1.0}{
\begin{tabular}{l | c c c}
\hline
\textbf{Method} & \textbf{CLD} & \textbf{CBS} & \textbf{CLD + CBS} \\
\hline
\textbf{remained annotations (\%)} & 99.60 & 4.51 & 4.10 \\
\hline
\end{tabular}
}
\label{table_remain_annotation}
\end{table}

Similarly, we denote the final predictions generated by $\Psi_{cls, c}$ with all features as inputs as $\widetilde{y}_{a}$, with the predicted category of a prediction as $\hat{y}_{a}$ and the confidence score of a prediction as $\check{y}_{a}$. The total loss can be written as:
\begin{equation}\label{eq_sup_all}
    \mathcal{L}_{a} = \lambda_1\mathcal{L}_{c} + \lambda_2\mathcal{L}_{n}
\end{equation}
where $\lambda_1$ and $\lambda_2$ are the trade-off parameters.

It should also be noticed that the assistance of the noisy feature will also reduce the reliance of the clean model on the warm-up stage. It can be inferred from the CLD that if the warm-up stage lasts for too long, the model will fit the distribution of the noisy label and misclassify some noisy annotations as clean ones, which will lead to the model training being biased by the noisy labels. However, if the warm-up stage is too short, the model cannot learn the distribution of the data, thus the clean label disentangling is also not that accurate. Therefore, the CLD is sensitive to the warm-up stage. On the contrary, the assistance of the noisy feature will provide a lower bound of the clean model, and we only need to learn a coarse fusion layer and a clean classifier, which are quite light-weighted. Moreover, the fusion layer and the clean classifier can also be finely trained during the following training stage. Therefore, only a few epochs of warm-up can also lead to a great performance.

To sum up, our noisy feature-aided clean label disentangling (NA-CLD) framework can sample out clean labels to prevent the training from being disturbed by the noisy annotations. In addition, our NA-CLD framework can also reduce the influence of the class-imbalanced problem, leading to an unbiased training process. Moreover, our NA-CLD framework can also take full use of the given annotations to learn more reliable semantics for better recognition performance.
\section{Experiments}
\label{sec_exp}

\subsection{Datasets}
\label{sec_datasets}
We examine the effectiveness of our proposed NA-CLD method on the \textbf{G1020 Dataset} and the \textbf{dataset B}. The G1020 dataset is a large fundus dataset consisting of 1020 fundus images from 432 patients for glaucoma diagnosis, including 724 images belonging to healthy patients and 296 images belonging to glaucoma patients. All of the images are resized to a size of $512\times 512$. Most of the samples are with three annotations, \textit{i.e.}, the background region, the optic disk (OD) region and the optic cup (OC) region, while some of the samples are only with two annotations, \textit{i.e.}, the background region and the optic disk (OD) region, with the OC region missing~\cite{bajwa2020g1020}. 
The dataset B is a breast ultrasound image dataset consisting of 163 images, which includes 110 benign cases and 53 malignant cases. All of the images are resized to a size of $512\times 512$. All of the samples are with two annotations, \textit{i.e.}, the background region and the abnormal region~\cite{yap2017automated}.


\begin{table*}[t]\centering
\caption{Comparasion with the state-of-the-art methods. We conduct the experiments on the G1020 dataset with source-free domain adaptation (SFDA) noise and joint erode and dilation (JED) noise and report the IoU score (in \%) and Dice score (in \%) of the optic cup (OC), the optic disk (OD) and the average performance (Avg.).}
\setlength{\tabcolsep}{5mm}{
\begin{tabular}{clcccccc}
\hline
\multicolumn{1}{c|}{\multirow{2}{*}{Noise Type}}                                            & \multicolumn{1}{c|}{\multirow{2}{*}{Method}} & \multicolumn{2}{c|}{Optic Cup}                             & \multicolumn{2}{c|}{Optic Disk}                            & \multicolumn{2}{c}{Average}                 \\ \cline{3-8} 
\multicolumn{1}{c|}{}                                                                       & \multicolumn{1}{c|}{}                        & IoU                  & \multicolumn{1}{c|}{Dice}           & IoU                  & \multicolumn{1}{c|}{Dice}           & IoU                  & Dice                 \\ \hline
\multicolumn{1}{c|}{\multirow{11}{*}{\begin{tabular}[c]{@{}c@{}}SFDA\\ Noise\end{tabular}}} & \multicolumn{1}{l|}{Baseline}                & 47.64                & \multicolumn{1}{c|}{63.66}          & 59.29                & \multicolumn{1}{c|}{73.99}          & 53.43                & 68.82                \\
\multicolumn{1}{c|}{}                                                                       & \multicolumn{1}{l|}{Fully Supervised}             & 60.09                & \multicolumn{1}{c|}{74.70}          & 84.05                & \multicolumn{1}{c|}{91.28}          & 72.07                & 82.99                \\ \cline{2-8} 
\multicolumn{1}{c|}{}                                                                       & \multicolumn{1}{l|}{IoU}                     & 48.10                & \multicolumn{1}{c|}{64.33}          & 64.36                & \multicolumn{1}{c|}{77.89}          & 56.23                & 71.11                \\
\multicolumn{1}{c|}{}                                                                       & \multicolumn{1}{l|}{Dice}                    & 48.29                & \multicolumn{1}{c|}{64.60}          & 63.12                & \multicolumn{1}{c|}{77.05}          & 55.70                & 70.83                \\
\multicolumn{1}{c|}{}                                                                       & \multicolumn{1}{l|}{SCE}                     & 50.89                & \multicolumn{1}{c|}{66.99}          & 63.46                & \multicolumn{1}{c|}{77.29}          & 57.17                & 72.14                \\
\multicolumn{1}{c|}{}                                                                       & \multicolumn{1}{l|}{NR-Dice}                 & 48.42                & \multicolumn{1}{c|}{64.66}          & 63.21                & \multicolumn{1}{c|}{77.17}          & 55.82                & 70.91                \\
\multicolumn{1}{c|}{}                                                                       & \multicolumn{1}{l|}{VolMin}                  & 44.51                & \multicolumn{1}{c|}{60.81}          & 62.87                & \multicolumn{1}{c|}{76.78}          & 53.69                & 68.80                \\
\multicolumn{1}{c|}{}                                                                       & \multicolumn{1}{l|}{CSS}                     & 46.00                & \multicolumn{1}{c|}{61.95}          & 68.47                & \multicolumn{1}{c|}{81.04}          & 56.97                & 71.30                \\
\multicolumn{1}{c|}{}                                                                       & \multicolumn{1}{l|}{JCAS}                    & 42.21                & \multicolumn{1}{c|}{57.59}          & 59.82                & \multicolumn{1}{c|}{73.93}          & 51.01                & 65.76                \\
\multicolumn{1}{c|}{}                                                                        & \multicolumn{1}{l|}{SC}                      & 47.69                & \multicolumn{1}{c|}{63.99}          & 61.24                & \multicolumn{1}{c|}{75.67}          & 54.46                & 69.83                \\
\multicolumn{1}{c|}{}                                                                       & \multicolumn{1}{l|}{NA-CLD (Ours)}           & \textbf{52.10}       & \multicolumn{1}{c|}{\textbf{68.32}} & \textbf{74.72}       & \multicolumn{1}{c|}{\textbf{85.43}} & \textbf{63.41}       & \textbf{76.87}       \\ \hline 
\multicolumn{1}{c|}{\multirow{11}{*}{\begin{tabular}[c]{@{}c@{}}JED\\ Noise\end{tabular}}}  & \multicolumn{1}{l|}{Baseline}                & 56.45                & \multicolumn{1}{c|}{71.43}          & 77.24                & \multicolumn{1}{c|}{87.09}          & 66.84                & 79.26                \\
\multicolumn{1}{c|}{}                                                                       & \multicolumn{1}{l|}{Fully Supervised}             & 60.09                & \multicolumn{1}{c|}{74.70}          & 84.05                & \multicolumn{1}{c|}{91.28}          & 72.07                & 82.99                \\ \cline{2-8} 
\multicolumn{1}{c|}{}                                                                       & \multicolumn{1}{l|}{IoU}                     & 55.04                & \multicolumn{1}{c|}{70.20}          & 82.10                & \multicolumn{1}{c|}{90.09}          & 68.57                & 80.15                \\
\multicolumn{1}{c|}{}                                                                       & \multicolumn{1}{l|}{Dice}                    & 53.63                & \multicolumn{1}{c|}{68.84}          & 81.80                & \multicolumn{1}{c|}{89.91}          & 67.71                & 79.37                \\
\multicolumn{1}{c|}{}                                                                       & \multicolumn{1}{l|}{SCE}                     & 56.15                & \multicolumn{1}{c|}{70.96}          & \textbf{82.75}       & \multicolumn{1}{c|}{\textbf{90.50}} & 69.45                & 80.73                \\
\multicolumn{1}{c|}{}                                                                       & \multicolumn{1}{l|}{NR-Dice}                 & 55.25                & \multicolumn{1}{c|}{70.57}          & 80.45                & \multicolumn{1}{c|}{89.07}          & 67.85                & 79.82                \\
\multicolumn{1}{c|}{}                                                                       & \multicolumn{1}{l|}{VolMin}                  & 53.87                & \multicolumn{1}{c|}{69.41}          & 80.00                & \multicolumn{1}{c|}{88.81}          & 66.94                & 79.11                \\
\multicolumn{1}{c|}{}                                                                       & \multicolumn{1}{l|}{CSS}                     & 57.55                & \multicolumn{1}{c|}{72.16}          & 80.75                & \multicolumn{1}{c|}{89.26}          & 69.15                & 80.71                \\
\multicolumn{1}{c|}{}                                                                       & \multicolumn{1}{l|}{JCAS}                    & 48.39                & \multicolumn{1}{c|}{64.12}          & 75.89                & \multicolumn{1}{c|}{86.04}          & 61.85                & 74.81                \\
\multicolumn{1}{c|}{}                                                                       & \multicolumn{1}{l|}{SC}                      & 52.19                & \multicolumn{1}{c|}{67.85}          & 80.34                & \multicolumn{1}{c|}{89.02}          & 66.26                & 78.42                \\
\multicolumn{1}{c|}{}                                                                       & \multicolumn{1}{l|}{NA-CLD (Ours)}           & \textbf{60.15}       & \multicolumn{1}{c|}{\textbf{74.83}} & 81.44                & \multicolumn{1}{c|}{89.37}          & \textbf{70.80}       & \textbf{82.28}       \\ \hline
\multicolumn{1}{l}{}                                                                        &                                              & \multicolumn{1}{l}{} & \multicolumn{1}{l}{}                & \multicolumn{1}{l}{} & \multicolumn{1}{l}{}                & \multicolumn{1}{l}{} & \multicolumn{1}{l}{} \\
\multicolumn{1}{l}{}                                                                        &                                              & \multicolumn{1}{l}{} & \multicolumn{1}{l}{}                & \multicolumn{1}{l}{} & \multicolumn{1}{l}{}                & \multicolumn{1}{l}{} & \multicolumn{1}{l}{} \\
\multicolumn{1}{l}{}                                                                        &                                              & \multicolumn{1}{l}{} & \multicolumn{1}{l}{}                & \multicolumn{1}{l}{} & \multicolumn{1}{l}{}                & \multicolumn{1}{l}{} & \multicolumn{1}{l}{}
\end{tabular}
\vspace{-30pt}
}   
\label{main_table_fundus}
\end{table*}

\begin{table}[t]\centering
\caption{Comparasion with the state-of-the-art methods. We conduct the experiments on the BUSI dataset with source-free domain adaptation (SFDA) noise and joint erode and dilation (JED) noise and report the IoU score (in \%) and Dice score (in \%).}
\setlength{\tabcolsep}{3mm}{
\begin{tabular}{clcc}
\hline
\multicolumn{1}{c|}{Noise Type}                                                             & \multicolumn{1}{c|}{Method}        & IoU                  & Dice                 \\ \hline
\multicolumn{1}{c|}{\multirow{11}{*}{\begin{tabular}[c]{@{}c@{}}SFDA\\ Noise\end{tabular}}} & \multicolumn{1}{l|}{Baseline}      & 54.22                & 69.60                \\
\multicolumn{1}{c|}{}                                                                       & \multicolumn{1}{l|}{Fully Supervised}   & 66.78                & 79.97                \\ \cline{2-4} 
\multicolumn{1}{c|}{}                                                                       & \multicolumn{1}{l|}{IoU}           & 54.32                & 69.58                \\
\multicolumn{1}{c|}{}                                                                       & \multicolumn{1}{l|}{Dice}          & 57.09                & 71.94                \\
\multicolumn{1}{c|}{}                                                                       & \multicolumn{1}{l|}{SCE}           & 58.57                & 72.87                \\
\multicolumn{1}{c|}{}                                                                       & \multicolumn{1}{l|}{NR-Dice}       & 56.28                & 71.66                \\
\multicolumn{1}{c|}{}                                                                       & \multicolumn{1}{l|}{VolMin}        & 53.98                & 69.05                \\
\multicolumn{1}{c|}{}                                                                       & \multicolumn{1}{l|}{CSS}           & 55.54                & 70.19                \\
\multicolumn{1}{c|}{}                                                                       & \multicolumn{1}{l|}{JCAS}          & 58.67                & 72.60                \\
\multicolumn{1}{c|}{}                                                                       & \multicolumn{1}{l|}{SC}            & 57.85                & 72.02                \\
\multicolumn{1}{c|}{}                                                                       & \multicolumn{1}{l|}{NA-CLD (Ours)} & \textbf{59.86}            & \textbf{74.38}            \\ \hline 
\multicolumn{1}{c|}{\multirow{11}{*}{\begin{tabular}[c]{@{}c@{}}JED\\ Noise\end{tabular}}}  & \multicolumn{1}{l|}{Baseline}      & 54.06                & 68.87                \\
\multicolumn{1}{c|}{}                                                                       & \multicolumn{1}{l|}{Fully Supervised}   & 66.78                & 79.97                \\ \cline{2-4} 
\multicolumn{1}{c|}{}                                                                       & \multicolumn{1}{l|}{IoU}           & 60.51                & 75.15                \\
\multicolumn{1}{c|}{}                                                                       & \multicolumn{1}{l|}{Dice}          & 59.24                & 73.63                \\
\multicolumn{1}{c|}{}                                                                       & \multicolumn{1}{l|}{SCE}           & 64.85                & 78.21                \\
\multicolumn{1}{c|}{}                                                                       & \multicolumn{1}{l|}{NR-Dice}       & 59.23                & 73.09                \\
\multicolumn{1}{c|}{}                                                                       & \multicolumn{1}{l|}{VolMin}        & 57.51                & 72.21                \\
\multicolumn{1}{c|}{}                                                                       & \multicolumn{1}{l|}{CSS}           & 62.52                & 76.08                \\
\multicolumn{1}{c|}{}                                                                       & \multicolumn{1}{l|}{JCAS}          & 64.39                & 77.98                \\
\multicolumn{1}{c|}{}                                                                       & \multicolumn{1}{l|}{SC}            & 57.71                & 72.37                \\
\multicolumn{1}{c|}{}                                                                       & \multicolumn{1}{l|}{NA-CLD (Ours)} & \textbf{69.73}       & \textbf{81.60}       \\ \hline
\multicolumn{1}{l}{}                                                                        &                                    & \multicolumn{1}{l}{} & \multicolumn{1}{l}{} \\
\multicolumn{1}{l}{}                                                                        &                                    & \multicolumn{1}{l}{} & \multicolumn{1}{l}{} \\
\multicolumn{1}{l}{}                                                                        &                                    & \multicolumn{1}{l}{} & \multicolumn{1}{l}{}
\end{tabular}
\vspace{-30pt}
}
\label{main_table_breast}
\end{table}

\subsection{Noise Patterns}
We adopt two widely used noise patterns that are close to real-world applications, namely the source-free domain adaptation (SFDA) noise and the joint erode and dilation (JED) noise. 
In particular, we obtain the SFDA noise labels on the G1020 dataset by running a model pre-trained on another widely used fundus dataset ORIGA dataset~\cite{zhang2010origa} to obtain the pseudo-labels.
Similarly, we obtain the SFDA noise labels on the dataset B by running a model pre-trained on another widely used breast ultrasound image dataset BUSI dataset~\cite{al2020dataset} to obtain the pseudo-labels.
Moreover, we obtain the JED noise by performing iterative random erosion or dilation on the coverage area of each category, where we set the erosion or dilation kernel as a random integer between 2 and 5, with an erosion or dilation iteration as a random integer between 3 and 8. 

\subsection{Implementation Details}
In this work, we use DeepLabv3+~\cite{chen2018encoder} as our segmentation model, which utilizes ResNet-50~\cite{he2016deep} pre-trained on ImageNet~\cite{deng2009imagenet} as the backbone. Our fusion layer $\Psi_{fuse}$ consists of a two-layer convolutional layer with a hidden dimension of 256 followed by the BatchNorm layer~\cite{ioffe2015batch} and the ReLU function. We use an SGD optimizer for our experiments with the initial learning rate set as 0.001 on both datasets. We trained our model for 50 epochs on both datasets with a poly-learning rate scheduler and the batch size is set as 6 on both datasets. We resize the input images to 512 $\times$ 512 and no strong data augmentations are used in this work. We set the number of the warm-up epochs $\tau$ as 10, the threshold for selecting confident samples $\gamma$ as 0.9, the default clean label selecting strategy by ``Strategy 2'', the class-balanced sampling times $\rho$ as 10, and the coefficients $\lambda_1$ and $\lambda_2$ as the default settings of 1.0 and 20.0, respectively, where we set the $\lambda_2$ as 20.0 mainly to enable the magnitudes of $\mathcal{L}_{c}$ and $\mathcal{L}_{n}$ to be consistent. It should be mentioned that the hyperparameters we set are consistent on all datasets, which verifies the robustness of our methods.

\subsection{Experimental Results}
In this section, we compare our NA-CLD method with recent methods for medical image segmentation with noisy labels, including the methods that proposed noise robust loss functions like the SCE loss or the NR-Dice loss, as well as the methods that proposed to tackle the noisy label issue by learning a noise transition matrix like VolMin or JCAS, and some other methods like CSS and SC. We also compare our NA-CLD method with other methods that aim at solving the class-imbalanced problem, including IoU loss and Dice loss. In addition, we also report the results by directly using the given noisy labels as supervision, denoted as the ``baseline'' method, we also include the results that using the clean labels as supervision, denoted as the ``Fully Supervised'' method. Here we adopt the IoU score (in \%) and the Dice score (in \%) to evaluate the performance of different methods on both the G1020 dataset and the BUSI dataset.

We first compare our methods with the others on the \textbf{G1020 dataset} and the results are reported in Table~\ref{main_table_fundus}. It can be inferred from the table that our NA-CLD achieves new state-of-the-art (SOTA) performance facing both the SFDA noise or the JED noise, exceeding the baseline method by nearly 10\% facing the SFDA noise and around 4\% facing the JED noise in terms of IoU score. Moreover, our method shows great potential in recognizing those hard-to-distinguish categories like the optic cup.
Our method also surpasses the previous SOTA method by a large margin of more than 6\% facing the SFDA noise and around 2\% fading the JED noise in terms of IoU score.
It should be noticed that our NA-CLD method can even outperform the fully supervised method facing the JED noise on the optic cup category, which is mainly due to our proposed class-balanced sampling strategy. 
It should also be mentioned that facing the datasets with severe class-imbalanced problems like the fundus datasets, the methods using the noise transition matrix seem to show poor performance. The main reason is that such methods can only estimate the class-wise transition relationship, while the class-imbalanced problem will severely influence such a relationship by revising the predictions to the commonly-seen categories, leading to poor performance. 
The pseudo-labeling methods like CS also cannot perform well, as the confirmation bias issue will mislead the training and thus degrade the performance. 
Under such a situation, those noise robust loss functions show their potential like the SCE loss.
Compared with these methods, our NA-CLD method can efficiently tackle the class-imbalanced problem using our newly proposed class-balanced sampling strategy, leading to excellent recognition performance.

We then compare our methods with the others on the \textbf{BUSI dataset} and the results are reported in Table~\ref{main_table_breast}. It can be inferred from the table that our NA-CLD achieves new state-of-the-art (SOTA) performance facing both the SFDA noise or the JED noise, exceeding the baseline method by a large margin. Note that our method exceeds the previous SOTA method by more than 1\% in terms of IoU score facing the SFDA noise and nearly 5\% in terms of IoU score facing the JED noise. It should especially be mentioned that our NA-CLD method surpasses the fully-supervised method (the ``Fully Supervised'') by a large margin of more than 2\% in terms of IoU score, indicating the effectiveness of our method.

\subsection{Ablation Studies}
In this section, we analyze the effectiveness of the detailed module designs of our NA-CLD approach on the G1020 dataset facing the SFDA noise. 

\textbf{Effectiveness of Different Components.} Recall that our NA-CLD method includes a clean label disentangling (CLD) framework, a class-balanced sampling (CBS) strategy and a noisy feature assistance (NFA) strategy. We now investigate the individual contributions of these modules in NA-CLD. The analysis results are reported in Table~\ref{ablation_module}. 

\begin{table}[t]\centering
\caption{Ablation study on the effectiveness of different components in our NA-CLD method, including a clean label disentangling (CLD) framework, a class-balanced sampling (CBS) strategy and a noisy feature assistance (NFA) strategy. We conduct the experiments on the G1020 dataset with SFDA noise and report the IoU score (in \%) of the optic cup (OC), the optic disk (OD) and the average performance (Avg.).}
\setlength{\tabcolsep}{3mm}{
\begin{tabular}{c c c | c c c}
\hline
CLD & CBS & NFA & OC & OD & Avg. \\
\hline
 & & & 47.64 & 59.23 & 53.43 \\
\checkmark & & & 43.28 & 58.61 & 50.94 \\
 & \checkmark & & 48.89 & 60.26 & 54.57 \\
\checkmark & \checkmark & & 50.38 & 73.92 & 62.15 \\
\checkmark & \checkmark & \checkmark & \textbf{52.10} & \textbf{74.72} & \textbf{63.41} \\
\hline
\end{tabular}
}
\label{ablation_module}
\end{table}

It can be inferred from the table that if we apply our CLD strategy only, the recognition performance of the model drops rapidly, the main reason is that the severe class-imbalanced problem will lead the model to mistakenly predict some rarely-seen classes, like the optic cup, to the commonly-seen classes, like the background. In consequence, the prediction score of those rarely-seen classes is also relatively low. When applying our CLD strategy, most of the rarely-seen annotations will be filtered out, which will make the class-imbalanced problem even worse, thus degrading the recognition performance of the model.

On the contrary, when combining our CLD method with our CBS strategy, as our CBS strategy can efficiently tackle the class-imbalanced issue, our CBS strategy can greatly fill the gap of the CLD method, leading to an amazing performance improvement of nearly 10\% over the CLD method in terms of IoU score. 

In addition, there is also over 1\% performance improvement when applying our CBS strategy alone, as our CBS strategy will tackle the class-imbalanced problem, leading to an unbiased training process, thus the recognition performance of the model would be improved.

Moreover, our NFA strategy will also enable the model to learn better semantics from both clean supervision and noisy supervision. The quality of the fused feature would be enhanced with the assistance of the noisy feature. Therefore, there is a performance improvement of more than 1\%.

In the following, we will analyze the detailed effects of each component. As the CLD framework can only show its potential with the assistance of the CBS strategy, we first discuss the effectiveness of our CBS strategy and then analyze the effectiveness of the CLD framework, finally, we will study the effectiveness of the NFA strategy.

\textbf{Effectiveness of CBS.} We first discuss the effectiveness of our CBS strategy, as the CBS strategy mainly tackles the class-imbalanced problem, we thereby analyze the sensitivity of the class-balanced sampling times $\rho$ and the results are reported in Table~\ref{ablation_CBS}. It can be inferred from the table that if $\rho$ is too small, \textit{e.g.}, 5, too many pixels will be discarded, which will cause a lot of information loss, resulting in a poor recognition result. On the other hand, if $\rho$ is too large, \textit{e.g.}, 20, the model would suffer from the class-imbalanced problem, leading to a biased training process, thus the recognition will also be influenced. It can be also inferred from the table that if $\rho$ is a moderate value, \textit{e.g.}, 7, 10 or 15, the model will benefit from the CBS strategy and the variation of the performance is not large, indicating the stability of our CBS strategy. Moreover, the performance will be even better if we set $\rho$ as 7, which, however, is an odd value compared with 10. Therefore, in this work, we set $\rho$ as 10 by default.

\begin{table}[t]\centering
\caption{Ablation study on the sensitivity of the class-balanced sampling times $\rho$ in our class-balanced sampling (CBS) strategy. We conduct the experiments on the G1020 dataset with SFDA noise and report the IoU score (in \%) of the average performance (Avg.). ``X'' indicates the baseline method without using the CBS strategy.}
\setlength{\tabcolsep}{2mm}{
\begin{tabular}{ l | c c c c c c}
\hline
$\rho$ & 5 & 7 & 10 & 15 & 20 & X \\
\hline
Avg. & 54.55 & 55.07 & 54.57 & 54.37 & 53.68 & 53.43 \\
\hline
\end{tabular}
}
\label{ablation_CBS}
\end{table}

\textbf{Effectiveness of CLD.} Here we the effect of the CLD framework when using or without using the CBS strategy, and the results are listed in Table~\ref{ablation_CLD} and Table~\ref{ablation_CBS_CLD}. It can be inferred from Table~\ref{ablation_CLD} that if we apply our CLD framework alone without using the CBS strategy, the higher the threshold $\gamma$, the lower the recognition accuracy the model can achieve. The main reason is that influenced by the class-imbalanced problem, the model is prone to generate confident predictions for those pixels belonging to the major class, while the predictions of the pixels belonging to the minor class are usually accompanied by low confidence scores. Therefore, most of the annotations belonging to the minor categories might be filtered out due to the high threshold, which will exacerbate the class-imbalanced problem, leading to performance degradation. A lower confidence threshold can ease the problem, but the clean label disentanglement framework cannot show its potential.

On the contrary, when combining the CLD framework with our CBS strategy, it can be inferred from the tables that our CBS strategy can efficiently tackle the class-imbalanced problem, thus the CLD framework can eventually show its potential to enable the model to learn from unbiased clean annotations, thus boost the recognition. A higher threshold $\gamma$ will greatly filter out noisy annotations, enabling the model to learn from clean annotations, thus the recognition performance of the model will be improved. For example, when adopting Strategy 2 to select the clean labels, a high $\gamma$ like 0.9 enables the model to achieve more than 10\% performance improvement in terms of IoU score when recognizing the optic disk and around 7\% average performance improvement in terms of IoU score, indicating the effectiveness of our CLD framework. 

Moreover, it can also be inferred from Table~\ref{ablation_CBS_CLD} that Strategy 1 relies heavily on the prediction of the model, which may be misled by the confirmation bias issue. Therefore, the performance of Strategy 1 is not stable. Moreover, the selected annotations take only a very small part, while such a small number of annotations can hardly provide enough semantics for clean model learning, leading to the degradation of the recognition performance. However, our proposed Strategy 2 can efficiently tackle the problem, as our Strategy 2 filters clean labels according to the voting results of the model prediction and the given noisy label sets. Therefore, our Strategy 2 can filter out more clean labels than Strategy 1, and can also mitigate the negative influence of the confirmation bias issue, leading to the improvement of the recognition performance.

\begin{table}[t]\centering
\caption{Ablation study on our clean label disentangling (CLD) framework \textbf{without} using our class-balanced sampling (CBS) strategy. We compare the different disentangling strategies and vary the threshold for selecting confident predictions $\gamma$. We conduct the experiments on the G1020 dataset with SFDA noise and report the IoU score (in \%) of the optic cup (OC), the optic disk (OD) and the average performance (Avg.).}
\setlength{\tabcolsep}{4mm}{
\begin{tabular}{c|c|ccc}
\hline
Strategy                    & $\gamma$ & OC    & OD    & Avg.  \\ \hline
Baseline                    & -        & 47.64 & 59.23 & 53.43 \\ \hline
\multirow{6}{*}{Stategy 1}  & 0.4      & 46.85 & 60.63 & 53.74 \\
                            & 0.5      & 47.20 & 59.58 & 53.39 \\
                            & 0.6      & 45.32 & 58.07 & 51.69 \\
                            & 0.7      & 44.45 & 59.35 & 51.90 \\
                            & 0.8      & 44.40 & 58.57 & 51.49 \\
                            & 0.9      & 42.18 & 58.51 & 50.34 \\ \hline
\multirow{6}{*}{Strategy 2} & 0.4      & 45.83 & 60.91 & 53.37 \\
                            & 0.5      & 44.74 & 61.23 & 52.99 \\
                            & 0.6      & 45.46 & 59.91 & 52.68 \\
                            & 0.7      & 43.58 & 59.73 & 51.65 \\
                            & 0.8      & 44.40 & 58.97 & 51.69 \\
                            & 0.9      & 43.28 & 58.61 & 50.94 \\ \hline
\end{tabular}
}
\label{ablation_CLD}
\end{table}

\begin{table}[t]\centering
\caption{Ablation study on our clean label disentangling (CLD) framework \textbf{with} using our class-balanced sampling (CBS) strategy. We compare the different disentangling strategies and vary the threshold for selecting confident predictions $\gamma$. We conduct the experiments on the G1020 dataset with SFDA noise and report the IoU score (in \%) of the optic cup (OC), the optic disk (OD) and the average performance (Avg.).}
\setlength{\tabcolsep}{4mm}{
\begin{tabular}{c|c|ccc}
\hline
Strategy                    & $\gamma$ & OC    & OD    & Avg.  \\ \hline
Baseline                    & -        & 47.64 & 59.23 & 53.43 \\ \hline
\multirow{6}{*}{Stategy 1}  & 0.4      & 49.50 & 61.07 & 55.29 \\
                            & 0.5      & 48.53 & 58.64 & 53.59 \\
                            & 0.6      & 48.19 & 61.38 & 54.78 \\
                            & 0.7      & 48.51 & 60.66 & 54.58 \\
                            & 0.8      & 47.56 & 60.00 & 53.78 \\
                            & 0.9      & 45.18 & 52.38 & 48.78 \\ \hline
\multirow{6}{*}{Strategy 2} & 0.4      & 48.73 & 57.97 & 53.35 \\
                            & 0.5      & 49.32 & 61.05 & 55.18 \\
                            & 0.6      & 48.97 & 63.49 & 56.23 \\
                            & 0.7      & 48.84 & 71.86 & 60.35 \\
                            & 0.8      & 49.16 & 71.96 & 60.56 \\
                            & 0.9      & 50.38 & 73.92 & 62.15 \\ \hline
\end{tabular}
}
\vspace{-3mm}
\label{ablation_CBS_CLD}
\end{table}

\textbf{Effectiveness of NF-CLD.} We further analyze the effectiveness of our NF-CLD framework and the results are listed in Table~\ref{ablation_NFCLD}. It can be inferred from the table that our NF-CLD framework will enable the clean model to extract more reliable information that contains more useful information. Compared with our CLD (w/ CBS) method which will filter out too many annotations, our NF-CLD framework will enable the clean model to learn from all of the annotations to extract potential information. Therefore, the performance of the clean model will be boosted compared with the CLD (w/ CBS) method, with a performance improvement of more than 1\%. Moreover, it can also be inferred from the table that the gradient calculated by the clean model will also back-propagate through the noisy encoder, thus providing unbiased and clean supervision for the noisy encoder. Therefore, the quality of the learned noisy feature can also be improved, and the performance of the noisy model can also be improved, \textit{e.g.}, an over 5\% performance improvement over the baseline method. 

\begin{table}[t]\centering
\caption{Ablation study on the effectiveness of our noisy feature-aided clean label disentanglement (NF-CLD) framework. We conduct the experiments on the G1020 dataset with SFDA noise and report the IoU score (in \%) of the optic cup (OC), the optic disk (OD) and the average performance (Avg.). We report both the performance of the clean model and the noisy model to verify the effectiveness of our NF-CLD framework.}
\setlength{\tabcolsep}{3mm}{
\begin{tabular}{ l | c c c c c c}
\hline
Method & OC & OD & Avg. \\
\hline
Baseline & 47.64 & 59.23 & 53.43 \\
CLD (w/ CBS) & 50.38 & 73.92 & 62.15 \\
NF-CLD (clean model) & 52.10 & 74.72 & 63.41 \\
NF-CLD (noisy model) & 51.73 & 65.53 & 58.63 \\
\hline
\end{tabular}
}
\vspace{-3mm}
\label{ablation_NFCLD}
\end{table}

\begin{table}[b]\centering
\caption{Ablation study on the sensitivity of the warm-up epochs $\tau$ in our CLD (w/ CBS) method and our NF-CLD method. We conduct the experiments on the G1020 dataset with SFDA noise and report the IoU score (in \%) of the average performance (Avg.).}
\setlength{\tabcolsep}{1mm}{
\begin{tabular}{ l | c c c c c c}
\hline
$\tau$ & 1 & 3 & 5 & 10 & 15 & 20 \\
\hline
CLD (w/ CBS) & 36.52 & 36.69 & 44.18 & 62.15 & 61.72 & 61.22 \\
NF-CLD & 63.10 & 62.87 & 62.93 & 63.41 & 63.03 & 63.00 \\
\hline
\end{tabular}
}
\label{ablation_warmup}
\end{table}

In addition, we also study the \textbf{influence of the warm-up}, and the results are reported in Table~\ref{ablation_warmup}. It can be inferred from the table that our CLD (w/ CBS) framework relies heavily on the warm-up stage, as the disentangling strategy of our CLD framework relies on the predictions of the model. If the warm-up stage lasts too long, the model will not only learn the distribution of the clean labels but also learn the distribution of the noise. Therefore, the model will produce incorrect predictions, which might be further selected as clean annotations, providing incorrect supervision, thus misleading the training and the degradation of the performance. If the warm-up stage lasts too short, the model can hardly learn the distribution of the clean labels, thus the generated predictions of the model would also be unreliable, thus the training might be influenced by the confirmation bias issue, also leading to a degradation of the performance. However, our NF-CLD framework consists of two streams, where the noisy stream is supervised by the initial noisy label sets from the beginning to the end, consistently producing supervision. Although the supervision may be not accurate, it still provides a lower bound for the clean stream. As the filtered clean annotations will also enhance the noisy feature learning, this lower bound will be improved during the training process, thus improving the performance of the model. Moreover, the clean model will also learn unbiased information based on the assistance of the noisy feature, thus achieving better recognition performance.

\begin{table}[t]\centering
\caption{Ablation study on the sensitivity of the coefficients $\lambda_1$ and $\lambda_2$ used in our NF-CLD method by fixing the value of $\lambda_1$ as 1.0 and varying the value of $\lambda_2$. We conduct the experiments on the G1020 dataset with SFDA noise and report the IoU score (in \%) of the average performance (Avg.).}
\setlength{\tabcolsep}{1mm}{
\begin{tabular}{ l | c c c c c c c}
\hline
$\lambda_2$ & 0.5 & 1 & 3 & 5 & 10 & 15 & 20 \\
\hline
Avg. & 62.74 & 63.01 & 62.72 & 63.10 & 63.28 & 63.20 & 63.41 \\
\hline
\end{tabular}
}
\vspace{-3mm}
\label{abtaion_coefficient}
\end{table}

Finally, we will investigate the \textbf{sensitivity of the coefficients} $\lambda_1$ and $\lambda_2$ used in our NF-CLD framework, and the results are reported in Table~\ref{abtaion_coefficient}. It can be inferred from the table that our method is robust to the coefficients, while our method achieves the best performance when $\lambda_2$ equals 20.0, where the magnitude of the two losses can be consistent.

\subsection{Visualization}
\begin{figure}[h]
\centering
\includegraphics[width=0.9\linewidth]{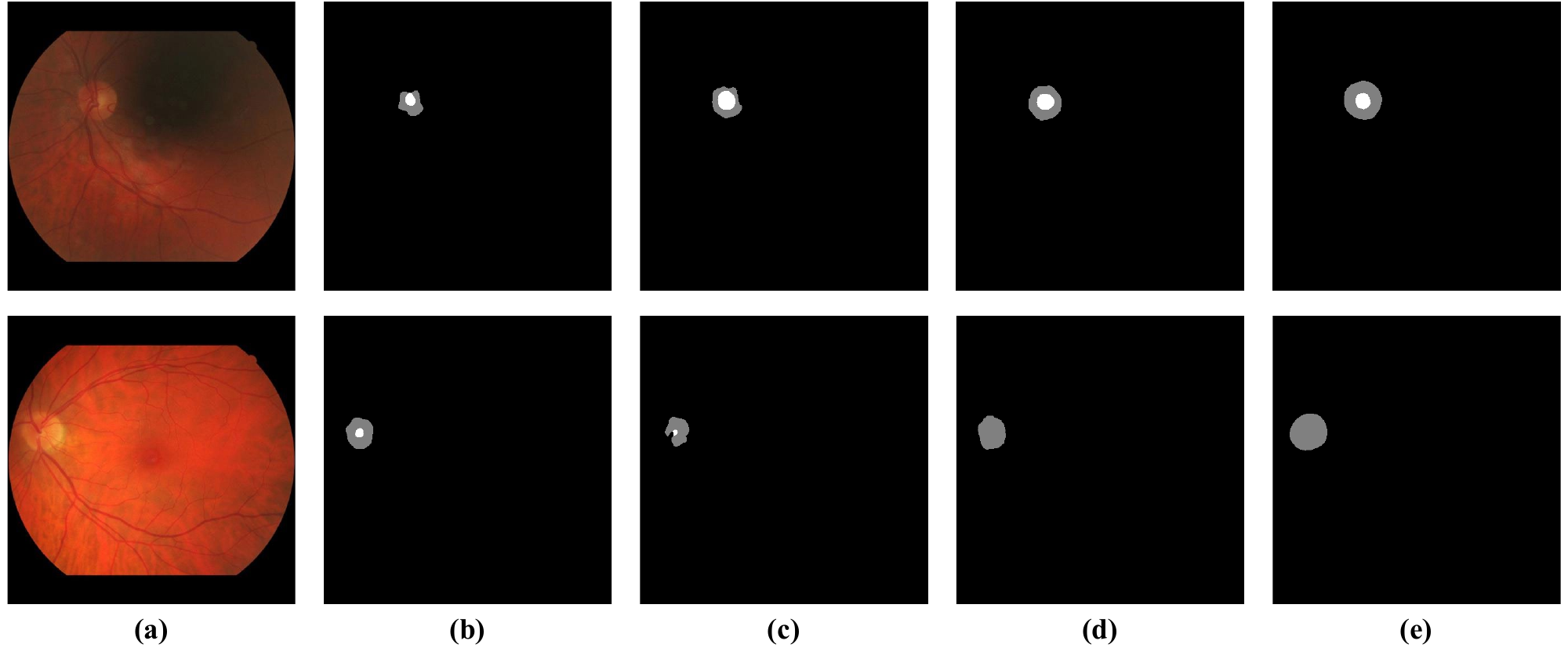}
\caption{Qualitative results of different methods on the G1020 dataset facing the SFDA noise. (a) input images, (b) the results of the SCE method, (c) the results of the SC method, (d) the results of our NA-CLD method, (e) the ground-truth labels.}
\label{fig_diff_method}
\end{figure}

In this section, we show some qualitative results of our NA-CLD method compared with previous state-of-the-art (SOTA) methods, including the best-performing method SCE and the up-to-date SC method, as shown in Fig.~\ref{fig_diff_method}. Note that we also list the input images and the clean ground-truth labels for comparison. We can infer from the figures that our method can achieve great recognition performance even facing some images with low quality (the first row), while the other methods may fail to recognize the pixels belonging to the optic disk. Moreover, when recognizing some fundus images belonging to glaucoma patients with optic cup atrophy (the second row), other methods may misclassify some pixels belonging to the optic disk to optic cup, while our method can efficiently distinguish these pixels. The qualitative results have validated the effectiveness of our NA-CLD method.

\section{Limitations and Future Works}
Despite the excellent performance of our method, our method introduces more network parameters and leads to times of computing complexity, which incurs a heavy burden on hardware requirements. In addition, in this work, we perform random sampling as our class-balanced sampling strategy. However, such an operation treats each pixel equally, while some hard pixels may need to be paid more attention. It is worth investigating to design more sampling strategies for better noise-robust performance.
On the other hand, how to extract more reliable information from the noisy features and how to better utilize the noisy features are still worth exploring. In this work, we only use a simple concatenation operation with two convolutional layers to fuse the noisy feature with the clean feature. However, such an operation is far from optimal, and it is possible to utilize the attention mechanism to assist the fusion. 
\section{Conclusion}
In this work, we first deeply analyze the current methods that tackle the medical image segmentation with noisy labels issue and then figure out the key issue that has been ignored by most works is the class-imbalanced problem. Then, we come up with a new clean label disentangling framework to encourage the model to learn from the selected clean and class-balanced annotations. We further extend our framework by proposing a noisy feature-aided clean label disentangling framework to enable the model to learn from both clean annotations and noisy annotations, as the noisy annotation may also contain valid semantics. Extensive experiments have validated the effectiveness of our newly proposed methods, where our methods achieve new state-of-the-art performance. 

{
    \small
    \bibliographystyle{ieeenat_fullname}
    \bibliography{main}
}


\end{document}